\documentclass[runningheads]{llncs}
\usepackage[T1]{fontenc}
\usepackage{graphicx}
\usepackage{multirow}
\usepackage{xcolor}
\usepackage{tikz}
\usepackage{placeins}
\usepackage[utf8]{inputenc}
\usepackage{amsmath}
\usepackage{todonotes}
\usepackage{algpseudocode}
\usepackage{algorithm}
\usepackage{caption}
\usepackage{framed}
\usepackage{float}
\usepackage{amssymb}
\usepackage{amsfonts}

\newcommand{\abele}{{\scshape abele}}
\newcommand{\poem}{{\scshape poem}}

\newcommand{\lore}{{\scshape lore}}

\begin{document}
\title{Explainable AI in Time-Sensitive Scenarios:\\Prefetched Offline Explanation Model}
\titlerunning{Explainable AI in Time-Sensitive Scenarios}
%
\author{Fabio Michele Russo\inst{1}\orcidID{0009-0003-4138-8822} \and
Carlo Metta\inst{2}\orcidID{0000-0002-9325-8232} \and
Anna Monreale\inst{3}\orcidID{0000-0001-8541-0284} \and
Salvatore Rinzivillo\inst{2}\orcidID{0000-0003-4404-4147} \and
Fabio Pinelli\inst{1}\orcidID{0000-0003-1058-6917}} 
\authorrunning{F. M. Russo et al.}
%
\institute{IMT School for Advanced Studies Lucca, Italy \and
ISTI-CNR, Italy \and University of Pisa, Italy}%
\maketitle              
\begin{abstract}
As predictive machine learning models become increasingly adopted and advanced, their role has evolved from merely predicting outcomes to actively shaping them. This evolution has underscored the importance of Trustworthy AI, highlighting the necessity to extend our focus beyond mere accuracy and toward a comprehensive understanding of these models' behaviors within the specific contexts of their applications. To further progress in explainability, we introduce \poem{}, Prefetched Offline Explanation Model, a model-agnostic, local explainability algorithm for image data. The algorithm generates exemplars, counterexemplars and saliency maps to provide quick and effective explanations suitable for time-sensitive scenarios. Leveraging an existing local algorithm, \poem{} infers factual and counterfactual rules from data to create illustrative examples and opposite scenarios with an enhanced stability by design. A novel mechanism then matches incoming test points with an explanation base and produces diverse exemplars, informative saliency maps and believable counterexemplars. Experimental results indicate that \poem{} outperforms its predecessor \abele{} in speed and ability to generate more nuanced and varied exemplars alongside more insightful saliency maps and valuable counterexemplars.

\end{abstract}
\section{Introduction}

Explainable Artificial Intelligence (XAI) within the field of Machine Learning (ML) is gaining increased attention, underlining its critical importance for both the application and advancement of research in this domain. The pervasive deployment of ML across a wide range of automated systems demonstrates its significance. The pace at which ML is evolving in terms of capability and application scope is impressive. However, the adoption of ML models has not been without concerns. There is a growing discourse in both public and academic spheres about the ethical implications and the societal impact of deploying these AI-based technologies \cite{o2016weapons} \cite{barocas2019fairness}. Issues such as delegating decision-making processes to machines raise substantial ethical and fairness questions. Moreover, the trustworthiness of automated decision-making systems has been debated. Applications of ML that resulted in unfair or unethical outcomes have underscored the necessity for ML systems to be trustworthy, both in their methodology and application.
Introducing regulation into the AI field has provoked a mix of responses. There are concerns that regulations might restrain innovation or fail to achieve their intended goals, thus failing on ethical and technological fronts. The other perspective suggests that thoughtful regulation can prevent the most harmful uses of ML and guide research towards ethical applications \cite{european2019ethics}. Balancing the vast potential of ML with ethical and legal requirements, such as privacy, security, justice and fairness, is essential.
Indeed, the necessity for explanations in ML applications is crucial and particularly challenging in scenarios where \textit{time} is critical. Applications such as autonomous vehicle navigation, real-time medical diagnosis, and high-frequency trading demand accurate and immediate explanations of model decisions. 

The ability to provide real-time explanations ensures that autonomous systems are transparent and applicable when decisions need to be made quickly and efficiently. This requirement adds a layer of complexity to the development of explainable models, as they must be designed to operate under stringent time constraints without compromising the quality of explanations. Let's consider the case of a medical diagnosis system that must assist the doctor in the evaluation of a challenging patient's case. The system should provide a suggestion for the diagnosis and explain the reasons behind it timely. In this context, the system's ability to generate timely and accurate explanations is critical for the patient's well-being and the system's reliability.
Another example is the deployment of an AI system for monitoring production lines in a factory. In case an atypical condition is found, the system must be able to explain the reasons behind its decisions in real-time to allow the operators to intervene promptly in case of anomalies. 
A third scenario is the use of AI in the financial sector, where a timely explanation may be crucial to understand the reasons behind a sudden change in the market or to predict future trends.

This paper addresses the challenge of explaining black box image classifications in time-sensitive scenarios. Our work, leveraging an existing XAI method presented in~\cite{guidotti2019black}, aims to introduce a model-agnostic algorithm for generating timely explanations of image data. Compared to~\cite{guidotti2019black}, we severely reduce the time needed to generate explanations in the domain of image data and produce higher quality explanations. The implementation of our proposed method, along with the complete code, is publicly available\footnote{\url{github.com/gatto/poem}}, enabling full reproducibility.

The rest of this paper is organized as follows.
Section~\ref{sec:related} discusses related works.
In Section~\ref{ch:background} we introduce preliminary background on some aspects which are important to understand the details of our approach.
Section~\ref{ch:methodology} details \poem{} while Section~\ref{ch:experiments} presents the results of its experimental analysis.
Section~\ref{ch:conclusions} concludes the paper and identifies some future works.

\section{Related Work} \label{sec:related}


Research towards the creation of transparent AI systems has led to significant progress in understanding and explaining model decisions \cite{guidotti2019survey}. Despite their impressive performance, deep learning models, characterized by their \textit{black box} nature, present considerable challenges in transparency and accountability, motivating the development of XAI methods.

Visual explanations in image classification have drawn attention due to their intuitive appeal, with exemplars and counterexemplars offering straightforward insights into model reasoning by presenting similar or dissimilar training instances \cite{guidotti2019black} \cite{ISIC2021}. Beyond visual methods, local explainability techniques like LIME and SHAP provide insights into model decisions by perturbing input features and assessing the impact on output, facilitating a more detailed understanding of model behavior \cite{lore} \cite{ribeiro2016should} \cite{lundberg2017unified}.


The advancement of model-agnostic methods marks a significant milestone in XAI. This universality enhances the adaptability of XAI solutions, making them valuable tools for developers and researchers working with a diverse array of machine learning models \cite{guidotti2019survey}. In parallel, efforts in creating interpretable models from the outset, such as transparent neural networks and decision trees, aim to build systems whose operations can be inherently understood, thereby reducing the dependency on post-hoc explanation methods \cite{rudin2019stop}.
Furthermore, the development of synthetic instance generation through approaches like adversarial autoencoders, as seen in \abele{}, enriches the landscape of explainability by providing novel ways to explore model decisions in the latent space \cite{isic2023}.

Research in XAI also extends to the evaluation of explanations' effectiveness, where metrics and user studies assess how well explanations meet the needs of different stakeholders, including end-users, developers, and regulatory bodies \cite{hoffman2018metrics} \cite{bodria}. This evaluation is crucial in tailoring explanations to various audiences, ensuring they are technically accurate but also comprehensible and actionable.


Various methods have been developed in the literature to address the need for timely and accurate model explanations, a crucial challenge in time-sensitive scenarios. Approaches like real-time saliency mapping and dynamic feature attribution methods \cite{realtime_saliency} stand as notable competitors, offering insights into model decisions with minimal delay. Techniques such as incremental learning models \cite{incremental} have also been adapted to provide explanations on-the-fly. Despite the advancements offered by these methods, they often face trade-offs between the speed of explanation generation and the depth or quality of the insights provided. The real-time aspect may come at the cost of oversimplified explanations that might not fully capture the complexity of the decision-making process or require substantial computational resources, limiting their applicability.


\section{Background} \label{ch:background}

In this paper, we consider the problem of creating a model that produces on-demand informative explanations for image classification decisions by a black box. Given a black box $b$ and an image instance $x$ classified by $b$ with label $y$, i.e., $b(x)=y$, our goal is to efficiently create an explanation to clarify the internal functioning of the black box that contributed to the specific classification.
To this end, we exploit an existing explanation model, \abele{}~\cite{guidotti2019black}, to improve the generated explanations' quality and time efficiency.

\begin{figure}[t]
    \centering
    \includegraphics[width=0.7\columnwidth]{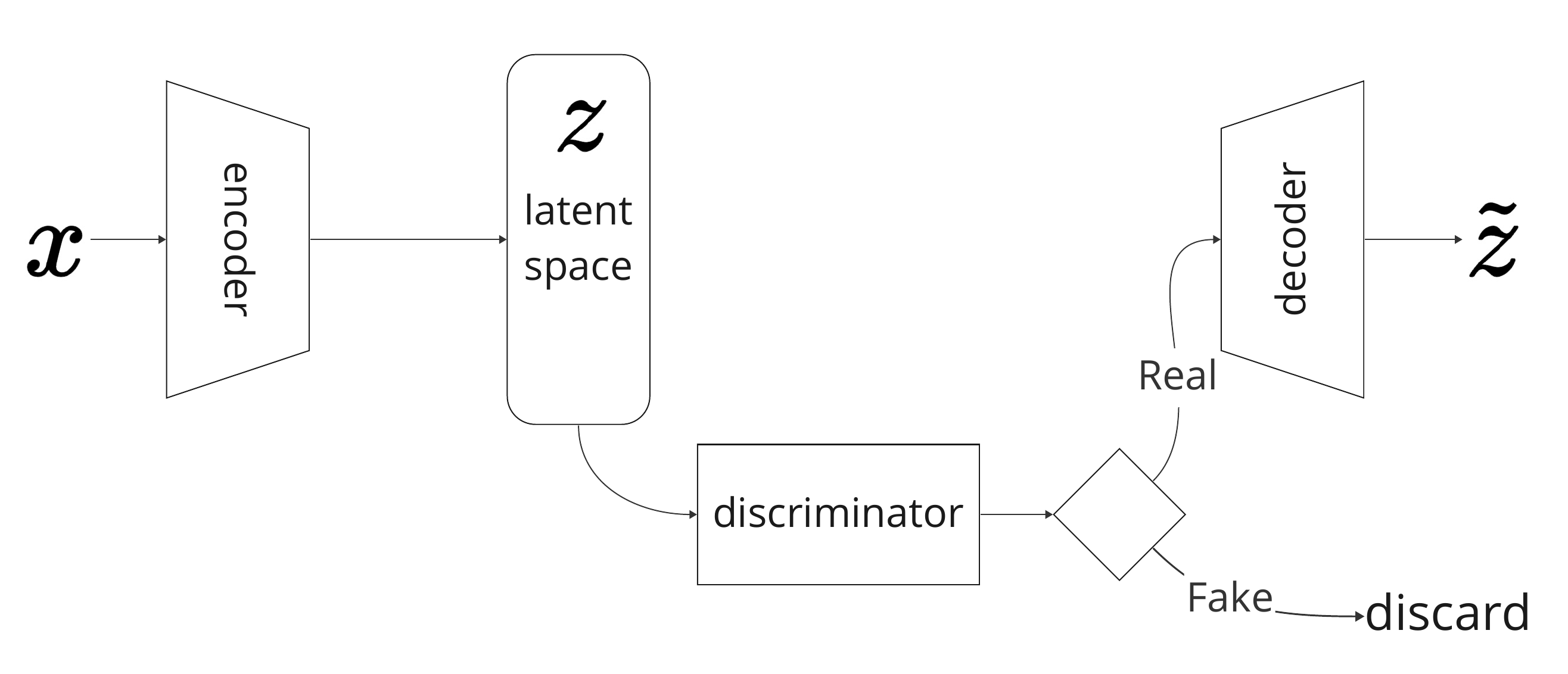}
    \caption{Components of the Adversarial AutoEncoder. The \textit{encoder} maps an input instance $x$ to a point $z$ into a latent space. Any point in this space is first filtered by the \textit{discriminator} before the \textit{decoder} reconstructs the corresponding image.}
    \label{fig:AAE}
\end{figure}

\abele{} is an explanation algorithm that follows the same strategy of \lore{}~\cite{lore}. Starting from an instance $x$, it generates a neighborhood of synthetic variations of $x$ and uses the black box to assign labels to this set. This annotated version of the neighborhood is used as a training sample for a decision tree. The logical predicates of the tree are used as source for factual and counterfactual rules. For the image domain, this process projects each instance into a latent space by means of an Adversarial Autoencoder (AAE)~\cite{makhzani2015adversarial} that encodes an image $x$ into a low-dimensional point $z$.
The AAE is also used to generate synthetic instances decoding one point $z$ from the latent space to a reconstructed image $\tilde{z}$. AAEs ensure synthetic instances retain the original data distribution through a structured framework involving an encoder, decoder, and discriminator, as detailed in Figure~\ref{fig:AAE}. The process aligns the latent space's aggregated posterior distribution with a specified prior, optimizing reconstruction accuracy\cite{makhzani2015adversarial}.

\begin{figure*}[tb!]
    \centering
    \includegraphics[width=\linewidth]{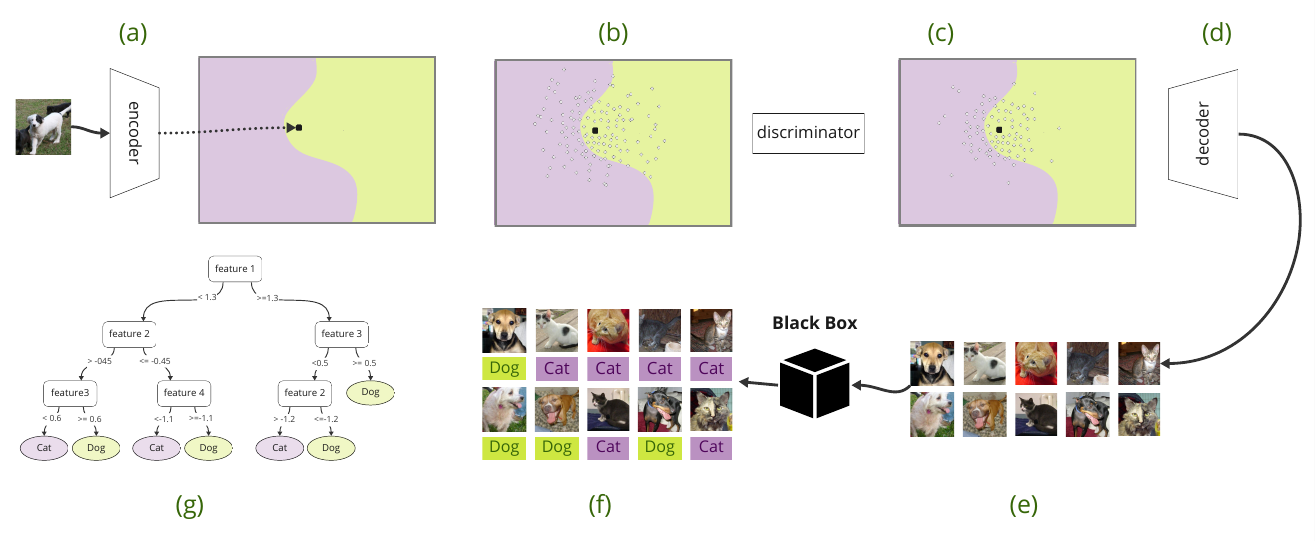}
    \caption{\abele{} workflow, starting from the mapping of the instance $x$ to the extraction of the transparent model: (a) instance $x$ encoded within the latent space; (b) neighborhood generation around $x$; (c) discriminator filtering the neighborhood; (d) decoder transforming the points into images (e) annotated by the black box; (f) supervised data forming a local training set to (g) learn a decision tree.}
    \label{fig:learning}
\end{figure*}

\textbf{Neighborhood Generation.}
\abele{} begins with the input $x$ and its class $y$ to encode the corresponding point $z$ in the latent space $Z$ (Fig.~\ref{fig:learning}(a)). Then, it generates a neighborhood $H$ around point $z$ (Fig.~\ref{fig:learning}(b)). Each element of $H$ is filtered by the discriminator (Fig.~\ref{fig:learning}(c)) and decoded to the corresponding image (Fig.~\ref{fig:learning}(d)). This set of reconstructed images is labeled by the black box $b$ (Fig.~\ref{fig:learning}(f)), and the labels are propagated back to the images' latent representation. \abele{} resamples points to guarantee a balanced distribution of the classes within $H$. Finally, a decision tree is trained on $H$, replicating the black box model's decisions locally to $z$ (Fig.~\ref{fig:learning}(g)).

\begin{figure}[h]
    \centering
    \includegraphics[width=0.7\columnwidth]{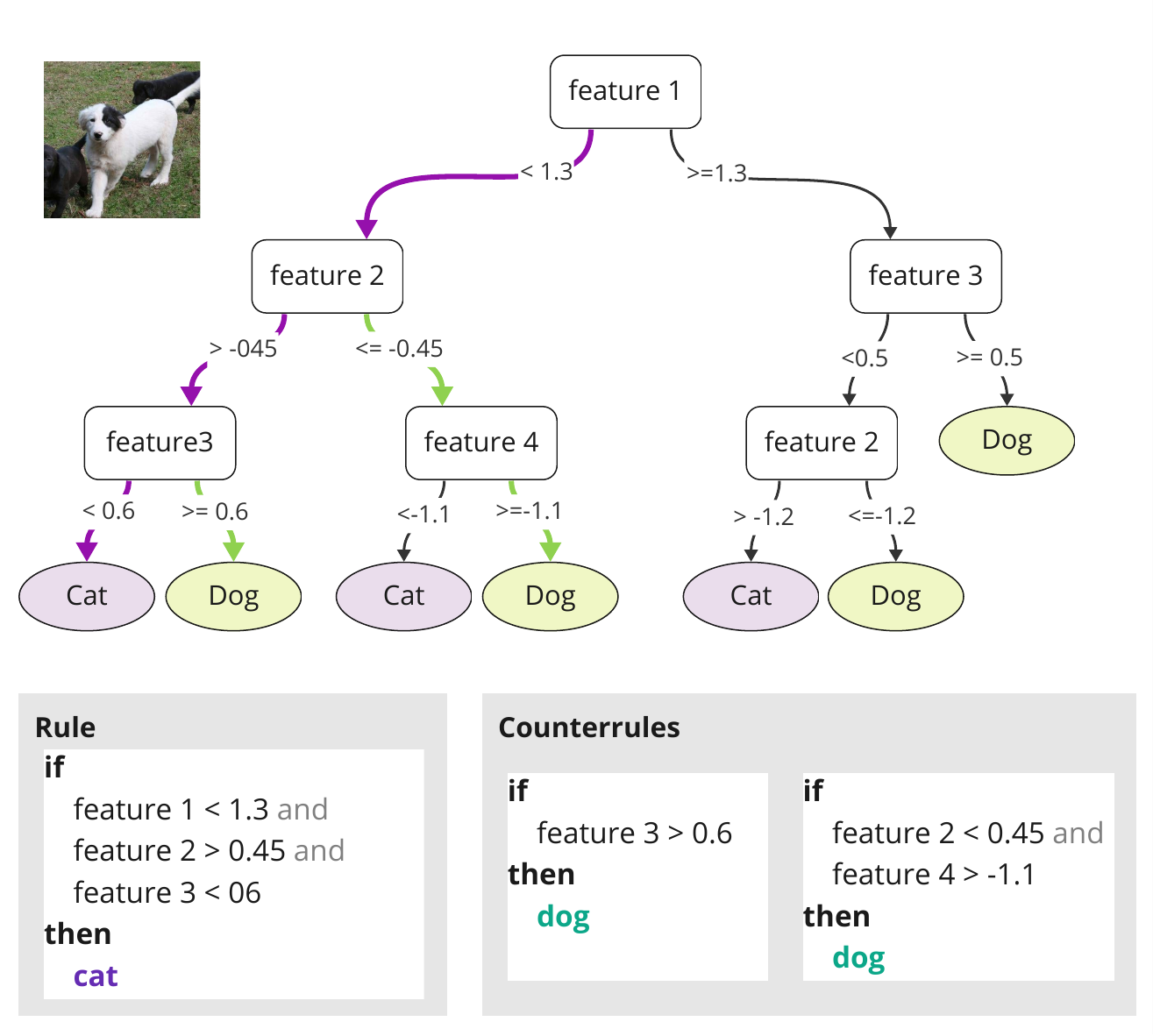}
    \caption{Extraction of rules and counter-rules from the surrogate model learned by \abele{} in the neighborhood $H$}
    \label{fig:surrogateModel}
\end{figure}

\textbf{Explanation Extraction.}
The decision tree classifies $z$, and the corresponding branch represents the factual rule $r = p \rightarrow b(x)$ for $z$.  In other words, the premise $p$  
is the conjunction of the splitting conditions in the nodes of the path from the
root to the leaf that is satisfied by the latent representation $z$ of the instance to
explain $x$ (Fig.~\ref{fig:surrogateModel}).


The counterfactual rules are determined by following alternative branches with minimal variations. In particular these rules have the form $q \rightarrow \neg b(x)$, where $q$ is the conjunction of splitting conditions
for a path from the root to the leaf labeling an instance $z'$ with $\neg b(x)$ and 
minimizing the number of splitting conditions falsified w.r.t. the premise $p$ of the rule $r$.




Rules are not directly usable for humans since they refer to latent dimensions that may not directly represent semantic attributes. Thus, to give the user intuition of the rule's meaning, we generate synthetic images as prototypes of the logical rules. \abele{} selects points in the latent space that satisfy the factual rule (or counterfactual rule) and pass the discriminator's filter. These points are decoded back and used as exemplars or counterexemplars for explanations.

\section{\poem{} - Prefetched Offline Explanation Model} \label{ch:methodology}

From the previous section, it is clear that producing an explanation from a given instance requires several steps for generating the neighborhood, decoding these points, classifying them, and learning the corresponding decision tree. This process may imply a large amount of time (see Table~\ref{tab:test-exectime}), which limits the actual use of the explanations, particularly in applications where interactivity or time sensitivity is crucial.
Therefore, we propose \poem{}, an explanation algorithm involving two steps called \textit{offline} and \textit{online}, respectively.

In the \textit{offline} step, \poem{} leverages \abele{} to build an \textit{explanation base} $\tilde{T}$, storing tuples which associate each explained point $t$ to the components crucial for its explanation $e(t)$ produced by \abele. 


In the \textit{online} step,  given an instance $x$, \poem{} selects from the explanation base $\tilde{T}$ the closest point to the latent representation of $x$ and then it uses the corresponding rules and counterfactual rules to generate exemplars and counterexemplars for $x$. 
 If the explanation base does not contains any tuple corresponding to a point sufficiently close to the point to be explained, the explanation is computed using \abele{}, and the result is stored in the explanation base $\tilde{T}$.

During the offline step, we assume to have enough time to dedicate to each explanation in $\tilde{T}$ to guarantee a higher quality of every single explanation.

\subsection{Offline step - Building the \textit{Explanation Base}}\label{sec:offline}
During this first phase, we assume the access to a sample of instances $D_e$ belonging to the same distribution of the training set of the black box $b$.
The dataset $D_e$ is sampled in two parts with respect to the class distribution: $D_a$, used to train the AAE, and $D_t$ used to build the explanation base $\tilde{T}$.

\begin{figure}[t]
    \centering
    \includegraphics[width=0.9\columnwidth]{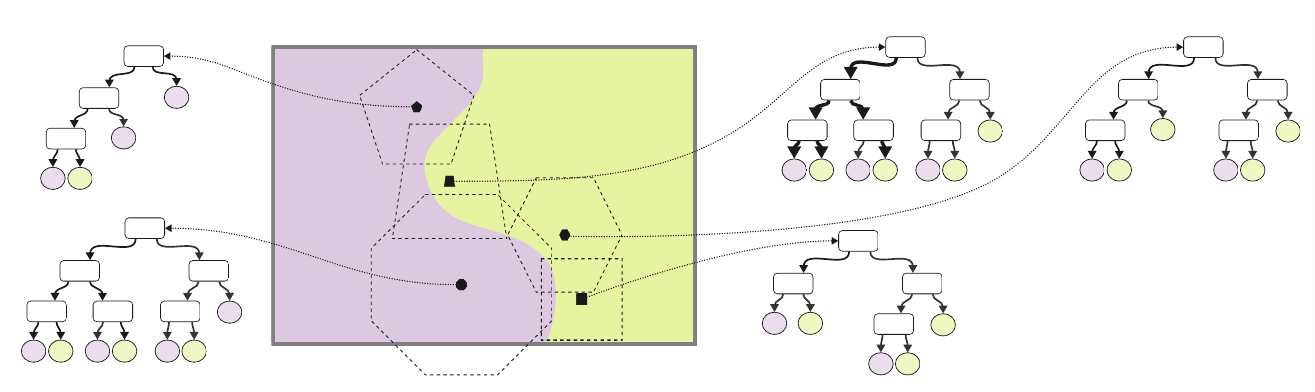}
    \caption{Representation of the explanation base, where each point $t \in D_t$ is linked to the decision tree and the neighborhood computed during the offline step.}
    \label{fig:explanationBase}
\end{figure}

For each point $t \in D_t$, we apply \abele{} to determine its corresponding explanation model. Thus we store the latent point $z_t$, the neighborhood $H_t$, and the surrogate decision tree $DT_t$ in a lookup index. In Figure~\ref{fig:explanationBase}, we show a schematic representation of the explanation base: each point $t$ is represented as a black dot, its neighborhood hyperpolyhedron as a dashed polygon and the corresponding decision tree is linked to each point in $D_t$. Each tuple of objects is stored, and an efficient indexing structure is used to retrieve the closest point to a given query.

\subsection{Online step -- Explanation of new instances}\label{sec:online}
Given a new instance $x$ to be explained, \poem{} exploits the explanation base to identify a pre-computed explanation model i.e, a pre-computed neighborhood and decision tree to generate the explanation. The instance $x$ is first mapped into a latent point $z_x$, then we look for the closest $z_t$ in the explanation base, using the Euclidean distance. We also check if all the following conditions hold:

\begin{enumerate}
\item $b$ classifies $x$ identically to the chosen $t$, i.e., $b(x)=b(t)$;
\item the latent representation $z_x$ is contained within the neighborhood hyperpolyhedron of the candidate point $z_t$;
\item the branch of the tree $DT_t$ that leads to the classification for $z_x$ is the same branch that holds for the explanation base point $z_t$.
\end{enumerate}

The first condition guarantees that we are using a pre-computed explanation model generated for explaining the classification of a point as class $b(x)$. This is important because we cannot assume a 100\% fidelity of the explanation model generated by \abele{}.  
The second condition ensures that the new point $z_x$ is part of the neighborhood of the previously computed neighborhood of $z_t$ in the explanation base. The third condition restricts the selection only on those surrogate models that share the same predicates on the new instance and the candidate point $z_t$. This also implies that the classification of $DT_t$ is the same for $z_x$ and $z_t$, i.e. $DT_t(z_x) = DT_t(z_t)$. We enforce this condition to guarantee that the explanation model is coherent with the new instance $x$. In future works, we plan to relax this condition to allow for a more flexible explanation model, even accounting for possible inconsistencies in the surrogate model.

If one of the conditions is not satisfied, the search proceeds by checking the second closest point $z_t'$, and so on until a valid candidate is found or no candidate may be selected.
When a candidate $t^* \in \tilde{T}$ is successfully identified, we build the rule and the counter-rules from the corresponding surrogate decision tree, using the same procedure described in Section~\ref{ch:background} and depicted in Figure \ref{fig:surrogateModel}. 
If no candidate is found, the explanation is computed using legacy \abele{} approach and the new explanation model is stored in the explanation base.

Once the suitable surrogate decision tree is identified and the rules and counter-factual rules are identified, \poem{} exploits them for generating exemplars and counter-exemplar images. We propose two novel strategies for generating such images differing from the \abele{} approach.

\subsubsection{Generation of Exemplars.}
\begin{figure}[ht]
    \centering
    \includegraphics[width=0.5\linewidth]{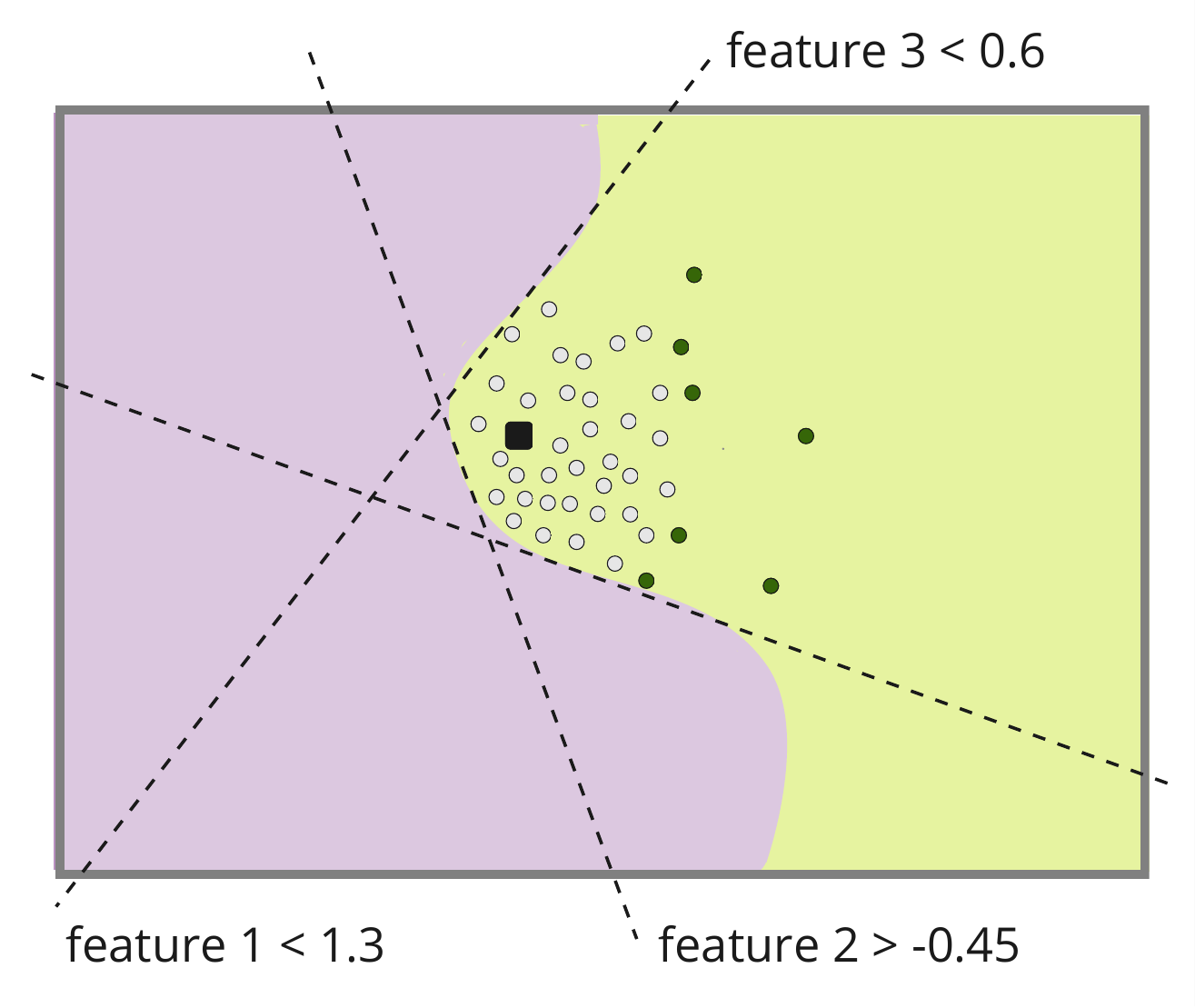}
    \caption{Generation and selection of exemplars for a point $z_x$. Dashed lines represent the half-planes determined by the predicates of the rule. A set of points is generated in the latent space and filtered by the rule predicate (gray points). The top $k$ distant points are selected as exemplars (green dots)}
    \label{fig:exemplar base}
\end{figure}

In this section, we describe the process for generating image exemplars. 
We assume that the user who needs an explanation for an image classification can request for a specific number $k$ of exemplars. To fulfill such request, \poem{} applies a procedure 
involving two phases: \textit{(i)} \textit{exemplar base generation}, aiming at generating a set of $\nu >> k$ exemplars; and \textit{(ii)} \textit{exemplar selection}, which selects the top $k$ most interesting exemplars among the generated ones.
\begin{description}
\item {\textit{Exemplar base generation:}} This step generates an \textit{exemplar base}, i.e., a set of $\nu$ exemplars where $\nu = k \times \beta$ ($\beta > 1$). 
In particular,  we generate exemplars by generating a new value for each latent feature.
Since the latent space learned by the AAE has dimensions that are distributed according to a standard normal distribution~\cite{makhzani2015adversarial}, we can generate a new value for each latent feature by drawing from a truncated normal distribution constrained to the feature conditions in the premises $p$ of factual rule extracted by the decision tree 
$r_{t^*} = p \rightarrow b(x)$.
For each feature $f_i$ involved in the premises $p$, we use the corresponding condition for bounding the feature values to be drawn from the standard normal distribution. 
As an example, if $p$ contains the condition $f_i < 0.5$ the value for the feature $f_i$ is drawn from the standard normal distribution in the interval $[-\infty, 0.5)$. 
For features which are not involved in the premises $p$ the value is drawn from the standard normal distribution without any bound. 
Once for each latent feature we generate a value, the obtained \textit{candidate exemplar} $c_e$ undergoes a discrimination check with the threshold set at some probability $\alpha$ and is classified by the back box $b$. The \textit{candidate exemplar} $c_e$ is stored in the exemplar base if its classification matches $x$'s (i.e., $b(c_e) = b(x)$) and it passes the discriminator check; otherwise, it is discarded.

\medskip

\item {\textit{Exemplar selection:}} This step extracts from the exemplar base the top $k$ most interesting exemplars. To this aim, we select the $k$ most distant exemplars from the point $x$ to be explained. The idea is to show the user exemplars having marked differences with respect to the image to be explained while maintaining the same classification label. In the latent space, this corresponds to selecting the exemplars that are farthest from the point $z_x$ to be explained.
To this end, \poem{} sorts the exemplar base with respect to the pairwise distance from $z_x$ by using the Euclidean distance. 
Then, the $k$ exemplars furthest from $x$ are presented to the user, where $k$ represents the number of requested exemplars. 
\end{description}
An illustrative depiction of the generated and selected exemplars from the \textit{exemplars base} is provided in Figure~\ref{fig:exemplar base}. We highlight that the figure shows a projection of the features in a two-dimensional space.



\subsubsection{Generation of Counterexemplars.}

In this section, we describe the procedure for generating counterexemplars. It is based on applying to the latent instance $z_x$ a perturbation guided by the counter-factual rules. 
In particular, given a candidate $t^*$, for each counterfactual rule $cr \in \Phi _{t^*}$ extracted from $DT_{t^*}$, \poem{} generates a counterexemplar. Given a counterfactual rule $cr = q \rightarrow \neg b(x)$, the premise $q$ is a conjunction of predicates of the form $(f\ op\ v)$, where $op$ can be $<,  \leq, >,$ or $\geq$ and contains some predicates $P$ falsifying the premise of the factual rule $r_{t^*}$.

For each feature $f$ in a predicate $(f \ op\ v)$ of $P$, we perturb the value of $f$ assigning a new value by adding (subtracting) to $v$ the value $\epsilon \times i$ in case the comparison operator $op$ of the predicate is $>$ or $\geq$ ($<$ or $\leq$). We highlight that $i = 1, \ldots, m$ while $m$ and $\epsilon$ are user parameters. The parameter $\epsilon$ allows the user to define the granularity of the search: larger (smaller) values will produce a counter-exemplar more (less) quickly but with a coarser (finer) resolution.

The iteration over $i$ is necessary because adding $\epsilon$ to the initial feature value could generate a candidate $c$ that does not pass the discriminator check or is not classified by the black box with the same label as $x$. Consequently, we iterate on $i$, increasing step by step the perturbation value until we find the suitable candidate $c$ passing the discriminator check and for which $b(x) \neq b(c)$.
In cases where a valid counterexemplar is not produced after $m$ attempts, the counterfactual rule is discarded and does not produce any counterexemplars.

Considering for example an instance $x$ to be explained, mapped to a latent point  $z_x = [0.8, -0.34, 0.58, -0.13]$ and the relative counterfactual rule  $f_1 < 1.3 \land f_2 > - 0.45 \land f_3 \geq 0.60 \rightarrow \ \text{dog}$, \poem{} produces the counterexemplar $[0.8, -0.34, 0.64, -0.13]$ by adding $\epsilon = 0.04$ to the threshold value $0.60$ of the predicate and assigning the obtained value to $f_3$.

As a final note in this part, we remark that these two novel algorithms for exemplar and counterexemplar generation are independent from the rest of the work, i.e. the generation of the explanation base and the distinction in offline and online part. Therefore, the two algorithms could also be implemented directly within \abele{} to improve the quality of the generated explanations, without a significant increase in the computational cost, since \abele{} spends the most amount of its computational time on neighborhood generation.

\subsubsection{Saliency map generation.}

Our method, \poem{}, generates saliency maps in a very similar way to \abele{}. Both approaches aim to clarify why a black box model classifies images in certain ways using saliency maps. \abele{} creates these maps using exemplars from the latent feature space, highlighting important areas for classification. In \poem{}, we also use exemplars to create saliency maps, but our exemplars are produced differently. Although we follow the same steps as \abele{}, the novel way we generate our exemplars could lead to more detailed saliency maps. This means our maps might show different insights compared to those from \abele{}, even though the underlying construction process is the same.

\section{Experiments} \label{ch:experiments}
This section presents the empirical evaluation of \poem{}, with the objectives of assessing: \textit{i)} the impact of the \textit{explanation base} size on the model's performance; 
\textit{ii)} the model's time efficiency compared to the \abele{} algorithm; and \textit{iii)} the quality of generated saliency maps in regards to their capacity to identify the most important parts of the image correctly.

\subsection{Experimental Setting}
\textbf{Datasets.}
We conducted our experiments across three widely recognized datasets: MNIST is a dataset of handwritten grayscale digits, FASHION dataset is a collection of Zalando grayscale products (e.g. shirt, shoes, bag, etc.) and EMNIST dataset is a set of handwritten letters. MNIST and FASHION have 10 labels while EMNIST has 26 labels.
Details on the datasets are reported in Table \ref{tab:datasetbb}.

\begin{table}[t]
    \centering
    \caption{Datasets resolution, train and test dimensions, and AAEs reconstruction error regarding RMSE.}
    \begin{tabular}{|c||c|cc|cc|}
\hline
        & &  \multicolumn{2}{c|}{Size} & \multicolumn{2}{c|}{RMSE}\\
Dataset & Resolution  & Train & Test & Train & Test  \\
\hline
MNIST & $28\times28$ & $50k$ & $10k$ & $33.69$ & $40.73$  \\
FASHION & $28\times28$ & $50k$ & $10k$ & $28.25$ & $29.81$ \\
EMNIST & $28\times28$ & $131k$ & $14k$   & $35.41$ & $41.12$  \\
\hline
\end{tabular}

    \label{tab:datasetbb}
\end{table}


\textbf{Back Box Classifiers and AAE.}
For each dataset, we employed two black box classifiers: a Random Forest (\textit{RF}) and a Deep Neural Network (\textit{DNN}), following the original choice made in \abele{}~\cite{guidotti2019black}.
For the \textit{RF} classifier, we used an ensemble of 100 decision trees, the minimum number of samples in each leaf of 10, and using the Gini coefficient to measure the quality of splits. The \textit{DNN} architecture consisted of three convolutional layers followed by two fully connected layers, with ReLU activations. We used a dropout rate of 0.25 to prevent overfitting and trained the model using the Adam optimizer with a learning rate 0.001. This architecture was selected for its proven effectiveness in image classification tasks.

Overall, classifier performances were consistent with established benchmarks in literature: for both MNIST and FASHION datasets, \textit{RF} achieved accuracies around 95-96\%, whereas \textit{DNN} performed better with accuracies ranging in 98-99\%. For EMNIST dataset, \textit{RF} accuracy stood at approximately 90\%, while \textit{DNN} showed again enhanced performance, reaching around 95\%.

Regarding the adversarial autoencoder, we opted for a symmetrical encoder-decoder structure with a tower of 3 convolutional blocks from 64 to 16 channels, a latent space of 4 dimensions for MNIST and 8 for FASHION and EMNIST. The discriminator consists of two fully connected layers of 128 and 64 neurons. Batch Normalization and Dropout were used to improve training and performance. The autoencoders were trained for 10,000 steps with a batch size of 128 images.

We conducted experimental analysis to identify a suitable $\alpha$ value, i.e., the probability threshold for the generated image acceptance by the discriminator. To this end, we used MNIST. We observed that the mean discriminator probability for MNIST is 0.349, with a median of 0.310. We use the highest value between these two to estimate the discriminator's accuracy across the dataset. As these values are extracted from real data, a threshold of $\alpha = 0.35$ is used to decide the validity of generated points for all our experiments.


\begin{figure*}[t]
    \centering
    \includegraphics[width=.93\linewidth]{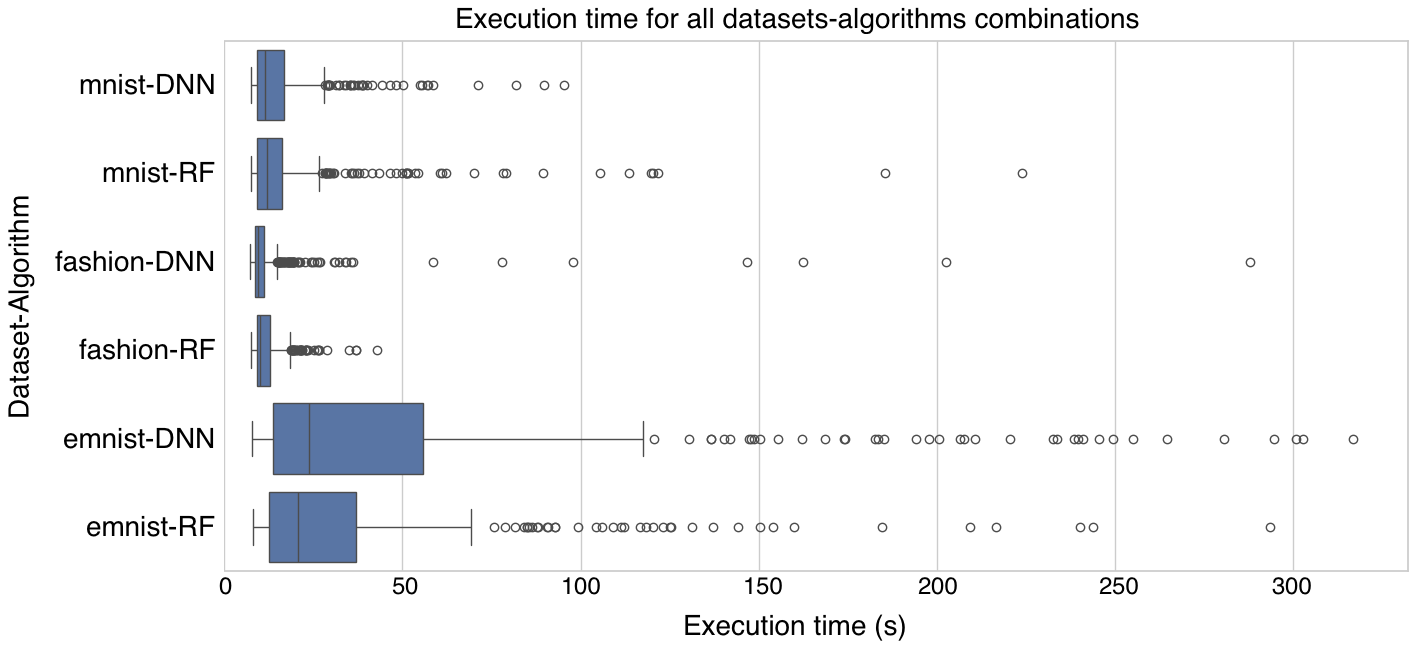}
    \caption{Time distributions for each dataset and black-box. For MNIST e FASHION, the majority of the instances get a successful hit in the explanation base, and they retrieve an explanation very efficiently. For EMNIST, the large number of class labels makes the search for candidate explanations more complex.
    }
    \label{fig:boxplot}
\end{figure*}
\textbf{Evaluation metrics and \poem{} parameters.}
To evaluate the impact of using the explanation base for explaining a classification, we define the measure \textit{hit percentage} as the proportion of explanation requests for which an appropriate pre-computed explanation model $\tilde{t^*}$ is found in the explanation base $\tilde{T}$. The scope of this measure is to assess the effectiveness of the explanation base in providing explanations for new instances, avoiding the need to compute them from scratch.

In order to extract examplars we set the parameter $\beta = 10$, and to extract counterexemplars, we chose the step size $\epsilon = 0.04$ and a maximum number of iterations $m = 40$. The latent space for the counterexemplars generation is explored over a standard normal distribution, and its $95th$ percentile falls between $-2$ and $2$. The step size of 0.04 consists of a 1\% of this interval. We chose 40 as the maximum number of iterations.

\subsection{Impact of Explanation Base Size}
The first experiment consists of an analysis of the size of the explanation base. \poem{} has a higher efficacy when, for the new instances to be explained, there is a high probability of finding a hit in the explanation base. Thus, the size of the explanation base, i.e., the amount of points used for pre-computing explanations, is crucial to maximize the probability of a hit. 
This size is strictly related to the dimensionality of the latent space or the number of distinct class labels. 
In this experiment, we address the question of what is the size of the explanation base $\tilde{T}$ that is suitable in a 10-label dataset to get a \textit{hit percentage} high enough to ensure that the vast majority of explanations are generated through \poem{}. Also, we want to study the execution time of \poem{} over different sizes of $\tilde{T}$. We run this experiment with two black boxes, testing an explanation base of increasing sizes of 500, 1000, 2500, and 5000 pre-computed explanations.

\begin{table}[t]
    \centering
    \caption{Performance comparison of \poem{} across different explanation base sizes for the MNIST dataset. The time is the average time in seconds for \poem{} to generate one explanation.}
    \label{tab:test-explbase-mnistrf}
    \begin{tabular}{|c||c|c|c|} \hline 
          \multirow{2}{*}{Black box} & \multirow{2}{*}{Size} & \multicolumn{2}{c|}{\poem{}} \\ \cline{3-4}
          & & Hit \% & Time (std dev) \\ \hline
          \multirow{4}{*}{RF}
          & 500 & 86.60 & 15.94 (14.73) \\ \cline{2-4}
          & 1000 & 92.20 & 16.58 (16.52) \\ \cline{2-4}
          & 2500 & 96.20 & 17.12 (19.15) \\ \cline{2-4}
          & 5000 & 98.00 & 17.07 (19.00) \\ \hline
          \multirow{4}{*}{DNN}
          & 500 & 88.00 & 12.98 (8.31) \\ \cline{2-4}
          & 1000 & 92.20 & 13.80 (14.40) \\ \cline{2-4}
          & 2500 & 94.40 & 13.62 (14.25) \\ \cline{2-4}
          & 5000 & 95.80 & 15.24 (10.63) \\ \hline
    \end{tabular}
\end{table}

Our results on MNIST dataset are reported in Table~\ref{tab:test-explbase-mnistrf}. We can observe that increasing the size of the explanation base improves, as expected, the explanation time and hit percentage.  
Since explanation bases with $5000$ pre-computed explanations report a good level of hit percentage, in the subsequent experiments, we use this value as explanation base size.

\subsection{Time performance Assessment} \label{ch:exp-time}

In this section, we present the results of the time performance of our approach. Table \ref{tab:test-exectime} summarises our findings. The results revealed significant enhancements when utilizing \poem{} compared to \abele{}. We observed reductions in execution time ranging from 83.00\% to 96.50\%, depending on the dataset and black box. 

\begin{table}[t]
    \centering
    \caption{Average execution times to generate a single explanation using both \poem{} and \abele{} and the hit percentage for \poem{}. The time measurements for \poem{} consider the online phase only, since the offline phase is performed once and does not depend on the instance to be explained.}
    \label{tab:test-exectime}
    \begin{tabular}{|c|c|c|c|c|} \hline 
         \multirow{2}{*}{Dataset} & \multirow{2}{*}{Black box} & \multicolumn{2}{c|}{\poem{}} & \multicolumn{1}{c|}{\abele{}} \\ \cline{3-5}
         & & Hit \% & Time (std dev) & Time (std dev) \\ \hline

         \multirow{2}{*}{MNIST} & RF & 98.00 & 17.07 (19.00) & 301.04 (161.64) \\ \cline{2-5}
         & DNN & 95.80 & 15.24 (10.63) & 230.06 (135.52) \\ \hline
         \multirow{2}{*}{FASHION} & RF & 99.40 & 11.72 (4.46) & 334.75 (159.18) \\ \cline{2-5}
         & DNN & 99.40 & 12.76 (18.85) & 281.77 (183.00) \\ \hline
         \multirow{2}{*}{EMNIST} & RF & 72.89 & 35.11 (40.23) & 333.56 (190.19) \\ \cline{2-5}
         & DNN & 72.22 & 50.75 (63.65) & 298.58 (103.25) \\ \hline
    \end{tabular}
\end{table}

We observe that using the same size of $\tilde{T}$ for a dataset containing 26 class labels instead of 10
leads to reduced performance in terms of both time and hit percentage.
This is due to the fact that the \textit{online} step of \poem{} for explaining a point $x$ only considers pre-computed explanations of points labelled the same as $x$ by the black box.
Thus, increasing the number of classes without increasing the explanation base size leads to an under-coverage of the latent space.
Lastly, since the standard deviation of completion time is relevant, 
we analyzed more in detail the distribution of execution time over the instances to be explained. The boxplots in Figure \ref{fig:boxplot} show that the interquartile range are very small for the three datasets. MNIST and FASHION leverage the coverage of the explanation base, having a very efficient time to find an explanation, since the time variability is very low. EMNIST instead requires more time to explore the explanation base, mainly due to the higher number of class labels.

\begin{figure*}[t]
    \centering
    \includegraphics[width=0.44\linewidth]{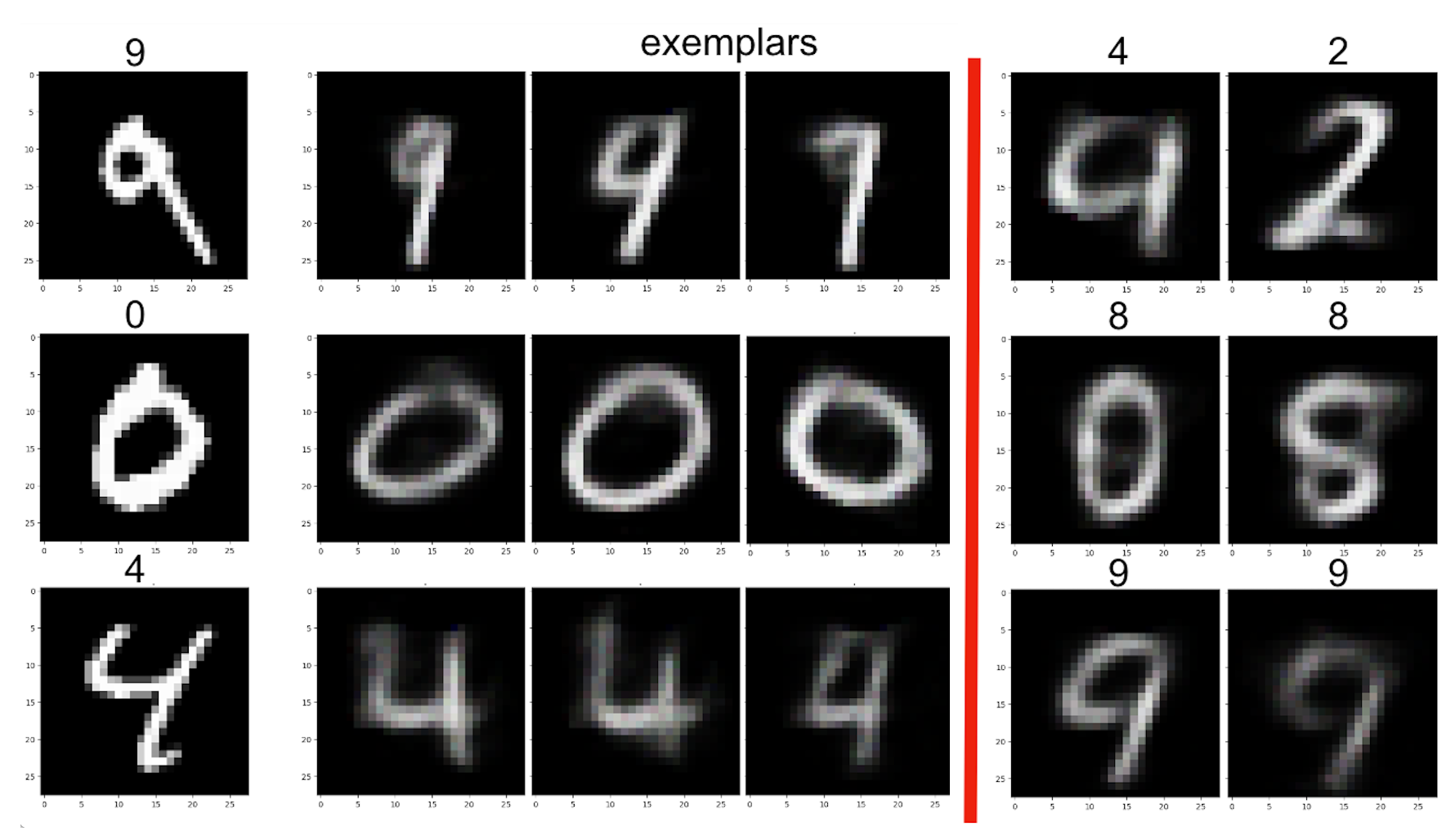} \quad
    \includegraphics[width=0.49\linewidth]{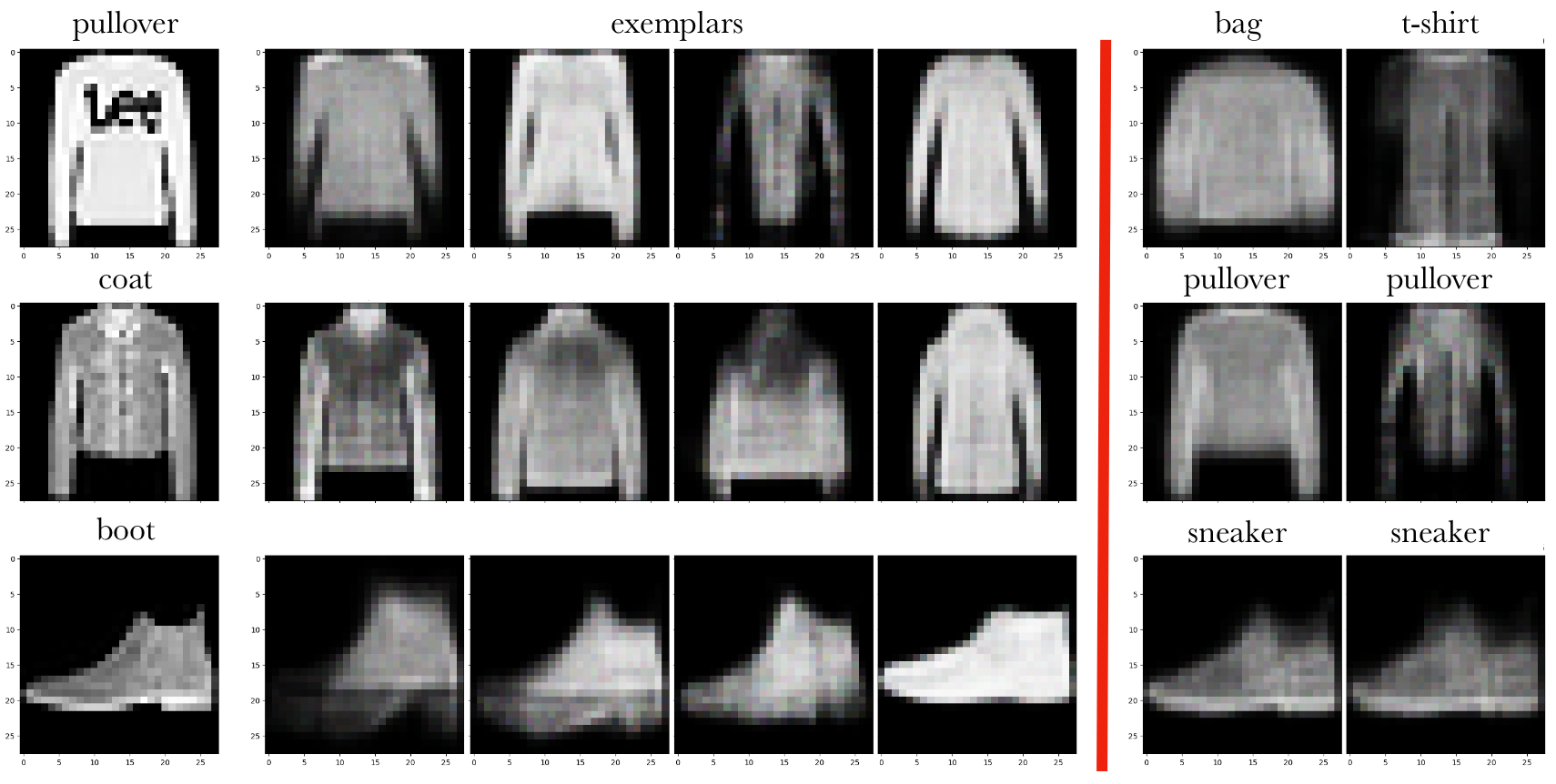}
    \caption{Three examples of explanations for instances of the MNIST (left) and FASHION (right) datasets. On the left, the image to be classified, with the class assigned by the black box on top. In the middle, a set of exemplars. On the right, two counter exemplars for each instance. }
    \label{fig:examples-mnist}
\end{figure*}

\subsection{Exemplars, Counterexemplars and Saliency Maps} \label{ch:exp-quality}

To demonstrate the efficacy of the approach in producing informative explanations, we performed two types of evaluation. The first is a qualitative comparison of the exemplars and counterexemplars produced by \poem{}. For example, in Fig. \ref{fig:examples-mnist}, we show a selection of images from MNIST and FASHION and their corresponding set of exemplars and counterexemplars. It is evident how the selection of the counterexemplars produces synthetic images that are visually similar to the original instance but with a different label. These counter exemplars push the black box to classify instances that are as close as possible to the decision boundary. This is evident in the second example of MNIST, where the counter exemplars are very similar to the original instance but with a different class label (apparently wrong for the human eye). The first instance of FASHION shows how one counter exemplar is labeled as a t-shirt, whereas at the human vision it resembles a pair of trousers.

\begin{figure}[bt]
    \centering
    \includegraphics[width=0.45\linewidth]{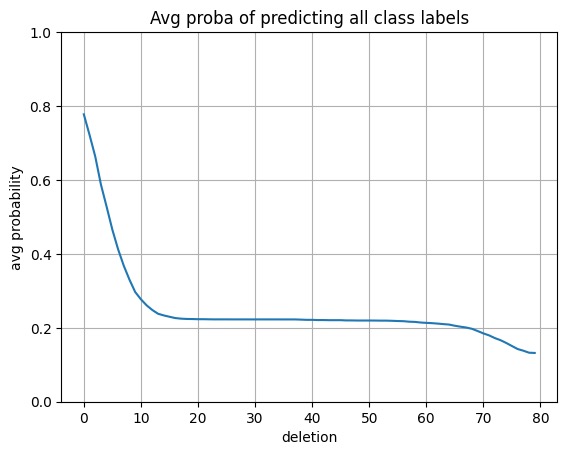}
    \caption{Deletion experiment over \poem{}'s saliency maps on a sample of 100 instances for each class label in the MNIST dataset. \textit{AUC}: 19.533.}
    \label{fig:test-deletion-average}
\end{figure}
\begin{figure}[t]
    \centering
    \includegraphics[width=.5\linewidth]{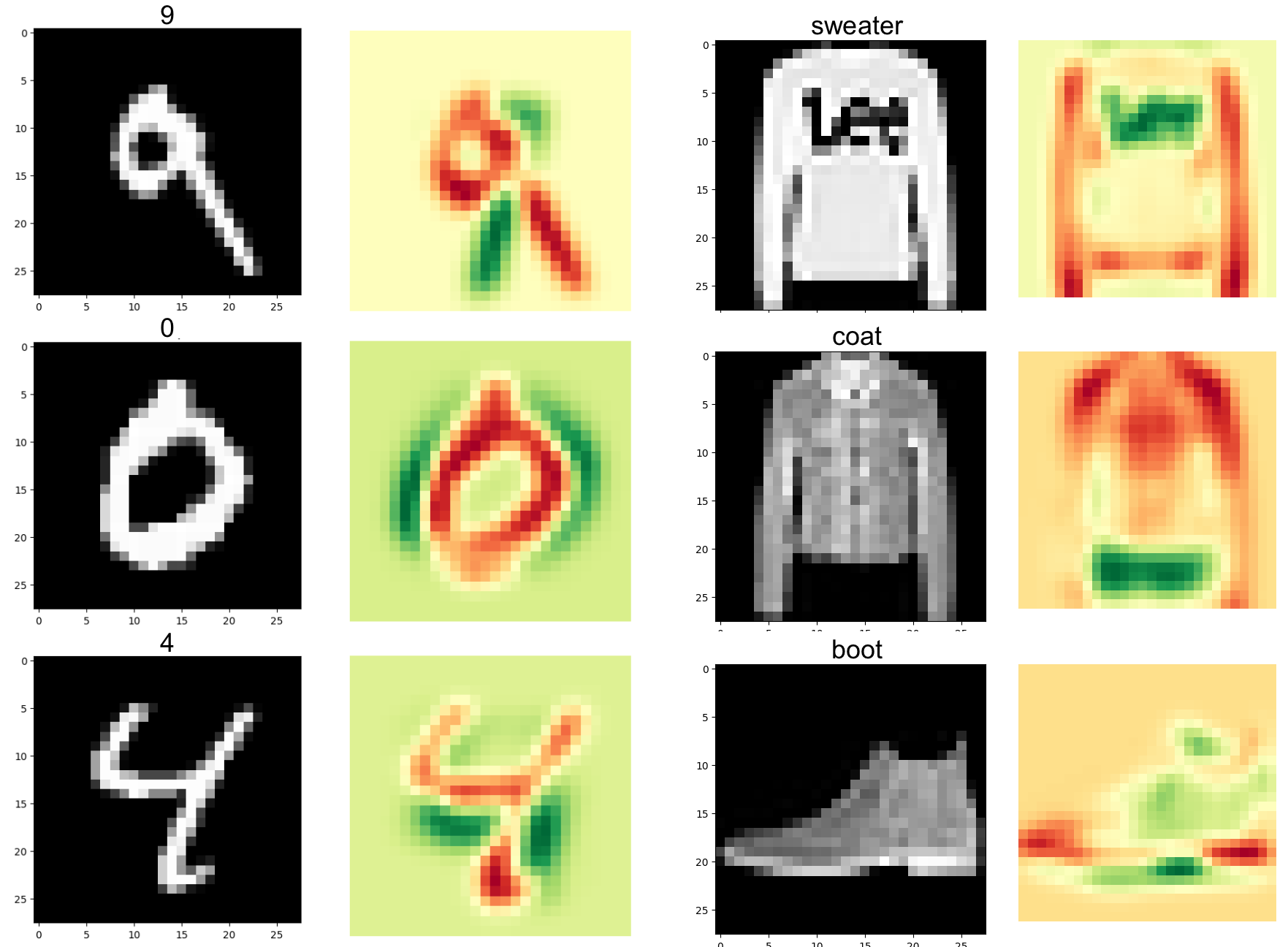}
    \caption{\poem{}'s saliency maps on MNIST and FASHION datasets. A divergent color scale maps the relevance of each pixel: red, positive contribution; green, negative contribution. The saliency maps highlight the most relevant parts of the image for the classification. For example, the classification of the sweater in the FASHION dataset is influenced by the presence of the sleeves and not by the brand writing.}
    \label{fig:examples-saliencies}
\end{figure}

To evaluate the efficacy of the saliency maps generated by \poem{}, we employed the deletion experiment method \cite{deletion}. This approach assesses the saliency map's capacity to determine each pixel's relevance. A steep decline in classification probability upon pixel removal, in a descending order on the blackbox's correct classification probability, implies a higher map quality. With pronounced drops observed in our deletion experiment curves, our results show that \poem{} consistently produces saliency maps that accurately identify important pixels for classification. In Figure~\ref{fig:test-deletion-average} it is evident how the deletion impacts the classification probability for a sample of 100 images. Figure \ref{fig:examples-saliencies} shows examples of saliency maps for a sample of images extracted from MNIST and FASHION dataset.

\section{Conclusions} \label{ch:conclusions}
In this paper, we show an algorithm called \poem{} for the model-agnostic explanation of black box decisions on image data that improves on the state-of-the-art of explanations through exemplars, counterexemplars and saliency maps. The proposed approach aims at speeding up the process of providing explanations to the final user while offering comprehensive explanations with good quality.  
By leveraging deterministic components and consistency checks, \poem{} ensures the selection of the same pre-computed explanation model for the same instance to be explained. Consequently, identical factual and counterfactual rules are applied across multiple runs, yielding highly stable explanations by design.
We tested our approach on benchmarking datasets, which resulted in a significant reduction of execution times. 
In future work, we intend to explore the possibility of extending this approach to different types of data, such as time series and tabular data. Moreover, it would be interesting to test the approach on real-world data, for example, in the field of medical images and to implement a user study for a qualitative assessment of the \poem{}'s explanations. Another interesting direction would be to investigate the possibility of exploiting the conditions selection of the online phase as an indicator for drifting in the data distribution along the use of the model, for example by triggering a model retraining when a higher number of missing conditions are detected.

\begin{credits}
\subsubsection{\ackname} Research partly funded by PNRR - M4C2 - Investimento 1.3, Partenariato Esteso PE00000013 - "FAIR - Future Artificial Intelligence Research" - Spoke 1 "Human-centered AI", funded by the European Commission under the NextGeneration EU programme, G.A. 871042 \emph{SoBigData++}, G.A. 101092749 \emph{CREXDATA},  ERC-2018-ADG G.A. 834756 \emph{XAI}, 
``SoBigData.it - Strengthening the Italian RI for Social Mining and Big Data Analytics'' - Prot. IR0000013, G.A. 101120763 \emph{TANGO}.

\subsubsection{\discintname} The authors have no competing interests to declare that are
relevant to the content of this article.
\end{credits}

\bibliographystyle{splncs04}
\bibliography{biblio}

\end{document}